%% file: main.tex
\documentclass[10pt,twocolumn,letterpaper]{article}

\usepackage{comment}
\usepackage[dvipsnames,table,xcdraw]{xcolor}
\usepackage{booktabs,multirow,float,graphicx,subcaption}
\usepackage{mathtools}
\usepackage{siunitx}
\usepackage[accsupp]{axessibility}

\usepackage[pagenumbers]{cvpr} % To force page numbers, e.g. for an arXiv version

\input{preamble}

\definecolor{cvprblue}{rgb}{0.21,0.49,0.74}
\definecolor{red}{rgb}{0.69, 0.0, 0.0}

\usepackage[pagebackref,breaklinks,colorlinks,allcolors=cvprblue]{hyperref}

\def\paperID{***} % *** Enter the Paper ID here
\def\confName{CVPR}
\def\confYear{2026}

\title{
\mymodel: Echocardiography Video Segmentation via \\ OrthogonalizedState Update and Anatomical Prior-aware Feature Enhancement
}
\input{sec/author}

\begin{document}

\maketitle
\input{sec/0_abstract}
                \input{fig/first/first}
\input{sec/1_intro}
                \input{fig/challenge/cha}
                \input{fig/Met/architecture}
\input{sec/2_0_related}
\input{sec/3_0_method}
    \input{sec/3_1_OSU}
                \input{tab/cap}
    \input{sec/3_2_APFE}
\input{sec/4_0_exp}
                \input{tab/type}
                \input{fig/Exp/vis_cap}
    \input{sec/4_2_0_com_sota}
                \input{tab/abl}
                \input{fig/abl/abl}
                
    \input{sec/4_3_0_abl}

\input{sec/5_limitations}
                \input{fig/fail/fail.tex}
\input{sec/6_conclusion}
\input{sec/ackn}

{
    \small
    \bibliographystyle{ieeenat_fullname}
    \bibliography{main}
}

% WARNING: do not forget to delete the supplementary pages from your submission 
\input{xsec/X_suppl}

% \clearpage
% \input{sec_re/re}

\end{document}

%% file: preamble.tex
\newcommand{\mymodel}{\textsc{OSA}\xspace}

\usepackage{xcolor}

\definecolor{osured}{rgb}{1.0, 0.6, 0.6}
\definecolor{osucyan}{rgb}{0.6, 1.0, 1.0}

\newcommand{\hlred}[1]{\colorbox{osured}{\textcolor{black}{#1}}}
\newcommand{\hlcyan}[1]{\colorbox{osucyan}{\textcolor{black}{#1}}}

\usepackage{microtype}

\usepackage{titletoc}

%% file: sec/author.tex
\author{
    Rui Wang\textsuperscript{1} \quad
    Huisi Wu\textsuperscript{1}\thanks{Corresponding author.} \quad
    Jing Qin\textsuperscript{2}
    \\
    {\small \textsuperscript{1}College of Computer Science and Software Engineering, Shenzhen University} \\
    {\small \textsuperscript{2}Centre for Smart Health, School of Nursing, The Hong Kong Polytechnic University} \\
    {\small \href{mailto:2400101058@mails.szu.edu.cn}{\nolinkurl{2400101058@mails.szu.edu.cn}}, \href{mailto:hswu@szu.edu.cn}{\nolinkurl{hswu@szu.edu.cn}}}
}

%% file: sec/0_abstract.tex
\begin{abstract}
Accurate and temporally consistent segmentation of the left ventricle from echocardiography videos is essential for estimating the ejection fraction and assessing cardiac function. However, modeling spatiotemporal dynamics remains difficult due to severe speckle noise and rapid non-rigid deformations. Existing linear recurrent models offer efficient in-context associative recall for temporal tracking, but rely on unconstrained state updates, which cause progressive singular value decay in the state matrix, a phenomenon known as rank collapse, resulting in anatomical details being overwhelmed by noise. To address this, we propose OSA, a framework that constrains the state evolution on the Stiefel manifold. We introduce the Orthogonalized State Update (OSU) mechanism, which formulates the memory evolution as Euclidean projected gradient descent on the Stiefel manifold to prevent rank collapse and maintain stable temporal transitions. Furthermore, an Anatomical Prior-aware Feature Enhancement module explicitly separates anatomical structures from speckle noise through a physics-driven process, providing the temporal tracker with noise-resilient structural cues. Comprehensive experiments on the CAMUS and EchoNet-Dynamic datasets show that OSA achieves state-of-the-art segmentation accuracy and temporal stability, while maintaining real-time inference efficiency for clinical deployment. Codes are available at \url{https://github.com/wangrui2025/OSA}.
\end{abstract}

%% file: fig/first/first.tex
\begin{figure}[t]
    \centering
        \includegraphics[width=0.98\linewidth]{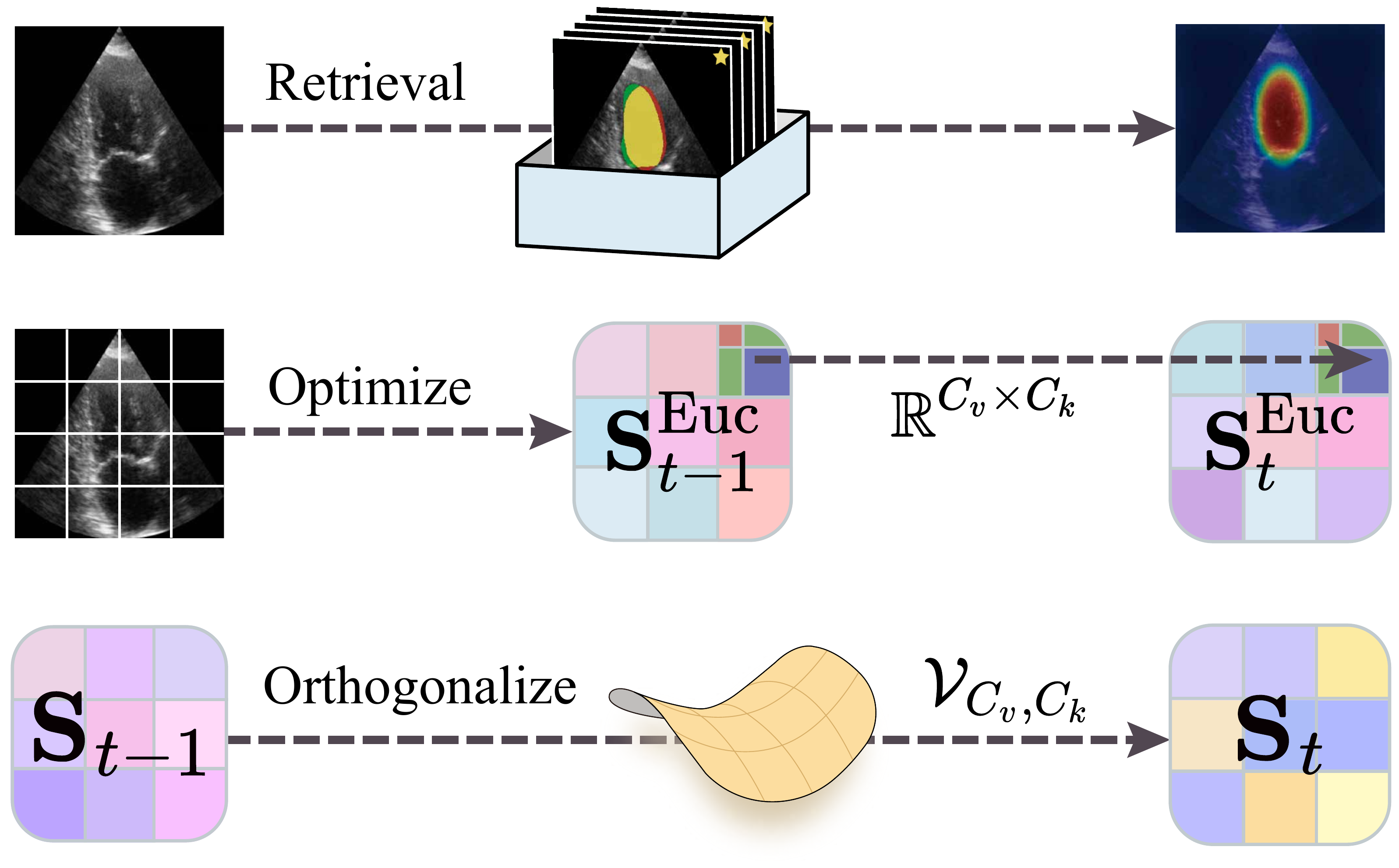}
    \caption{Illustration of different spatiotemporal memory update paradigms for echocardiography video segmentation. 
    \textbf{Top:} Memory bank methods rely on sparse key-frame retrieval to capture temporal dynamics. 
    \textbf{Middle:} Linear recurrent models perform unconstrained element-wise updates for state propagation. 
    \textbf{Bottom:} Our \mymodel enforces a Stiefel manifold constraint, yielding stable and drift-free tracking across the cardiac cycle.}
    \label{fig:first}
\end{figure}

%% file: sec/1_intro.tex
\section{Introduction}

\noindent
Echocardiography is a cornerstone of cardiovascular diagnosis due to its non-invasive, cost-effective, and real-time nature~\cite{10.3389/fcvm.2020.00025}.
Accurate segmentation of cardiac chambers, particularly the left ventricle (LV), from echocardiographic videos is essential for estimating the ejection fraction ($\text{LV}_{\text{EF}}$) and assessing precise cardiac function \cite{el2022semi,amer2021resdunet}.
Reliable and temporally consistent segmentation provides a foundation for quantitative evaluation of cardiac motion.
However, modeling the complex spatiotemporal dynamics of the heart under severe ultrasound noise remains a formidable challenge.
To address this, we propose a novel paradigm that fundamentally shifts from traditional frame-wise prediction or unconstrained state tracking to a manifold-constrained, matrix-level state evolution, as previewed in~\cref{fig:first}.

The difficulty of this task stems from complex imaging physics and rapid cardiac deformation. \Cref{fig:cha} shows that ultrasound imaging naturally suffers from severe speckle noise, low contrast, and limited spatial resolution, which obscure anatomical boundaries.
Meanwhile, the heart undergoes rapid non-rigid deformation between systole and diastole, causing pronounced appearance variations.
Furthermore, clinical datasets typically provide sparse annotations, such as exclusively at the end-diastole (ED) and end-systole (ES) frames, demanding models that can maintain temporal coherence and real-time efficiency under limited supervision.

Existing methods struggle to jointly satisfy these demands.
Early Convolutional Neural Networks \cite{MICCAI15_UNet,long2015fully} and Transformer-based models \cite{NIPS17_Attention,ICLR21_vit,TMI24_DSA} perform frame-wise prediction, largely ignoring continuous temporal dynamics while remaining computationally heavy.
Recent video modeling approaches adopt discrete memory retrieval mechanisms \cite{iccv19_stm,neurips21_stm,eccv22_xmem,iccv25_xmempp,cvpr24_cutie,ICLR25_SAM2,CVPR25_DAM,deng2024memsam,MICCAI25_SAMed2}, but their sparse key-frame updates struggle to fully utilize the complete historical information of videos (\cref{fig:first}, top).
Linear recurrent models (LRM) \cite{katharopoulos_transformers_2020,ICML20_ttt,ICML21_fwp,ICLR25_gdn,ICLR26_ttt_LaCT} offer efficient continuous tracking by compressing historical context into a fixed-size hidden state matrix $\mathbf{S}_t$.
Under sparse clinical annotations (\eg, only ED/ES frames), the model learns an in-context associative recall mechanism: during inference, it compresses historical observations into $\mathbf{S}_t$ to form implicit anatomical priors, then retrieves these priors using current noisy representations as queries.
However, existing LRMs rely on unconstrained Euclidean state updates.
Under severe speckle noise and rapid cardiac deformations, this lack of geometric regularization causes a progressive decay of singular values in $\mathbf{S}_t$.
This phenomenon, known as rank collapse, progressively degrades the model's associative memory capacity, severing the connection between current observations and historical priors, ultimately leading to long-term tracking inconsistency (\cref{fig:first}, middle).

Beyond the temporal rank collapse, the spatial domain presents its own persistent challenges.
Although recent methods \cite{deng2024memsam,mm25_echovim,ICCV25_gdkvm} have introduced architectural motifs tailored to ultrasound, these task-specific adaptations primarily focus on spatial feature re-weighting or specific frame sampling, leaving the underlying temporal propagation fundamentally unconstrained.
Without intrinsic structural regularities, the resulting features lack the necessary anatomical constancy required to distinguish true endocardial boundaries from transient acoustic artifacts.
This oversight leads to a systemic representation drift where anatomical information is gradually overwhelmed by stochastic speckle noise during long-sequence propagation.

These interwoven limitations motivate us to rethink both spatiotemporal state transitions and domain-specific feature extraction.
To address this, we propose \textbf{\mymodel}, a unified framework designed to jointly stabilize continuous temporal evolution and enhance anatomy-aware spatial representations.
By shifting from unconstrained linear recurrence to manifold-constrained nonlinear updates, and by explicitly embedding ultrasound-specific acoustic priors, \mymodel overcomes the core vulnerabilities of temporal rank-collapse and severe speckle interference.

Our main contributions are summarized as follows:
\begin{itemize}
    \item We propose \mymodel, a novel architecture designed for echocardiography video segmentation that successfully bridges the gap between temporal tracking stability and ultrasound-domain awareness, enabling robust, long-term cardiac modeling.
    \item We introduce the \emph{Orthogonalized State Update} (OSU) mechanism. OSU constrains the memory evolution to the Stiefel manifold via an orthogonalized update, effectively preventing rank collapse to ensure numerically stable temporal transitions and preserve complex structural details across the entire cardiac cycle.
    \item We design the \emph{Anatomical Prior-aware Feature Enhancement} (APFE) module. By reformulating spatial feature extraction as a physics-driven decoupling process, APFE explicitly separates persistent anatomical structures from stochastic speckle interference, providing the downstream temporal tracker with noise-resilient structural anchors.
    \item Comprehensive experiments on the CAMUS~\cite{tmi19_camus} and EchoNet-Dynamic~\cite{nature20_echodynamic} datasets demonstrate that \mymodel establishes a new state-of-the-art. It significantly improves both segmentation accuracy and temporal stability, while strictly maintaining the real-time inference efficiency required for clinical deployment.
\end{itemize}

%% file: fig/challenge/cha.tex
\begin{figure}[t]
    \centering
    \includegraphics[width=\linewidth]{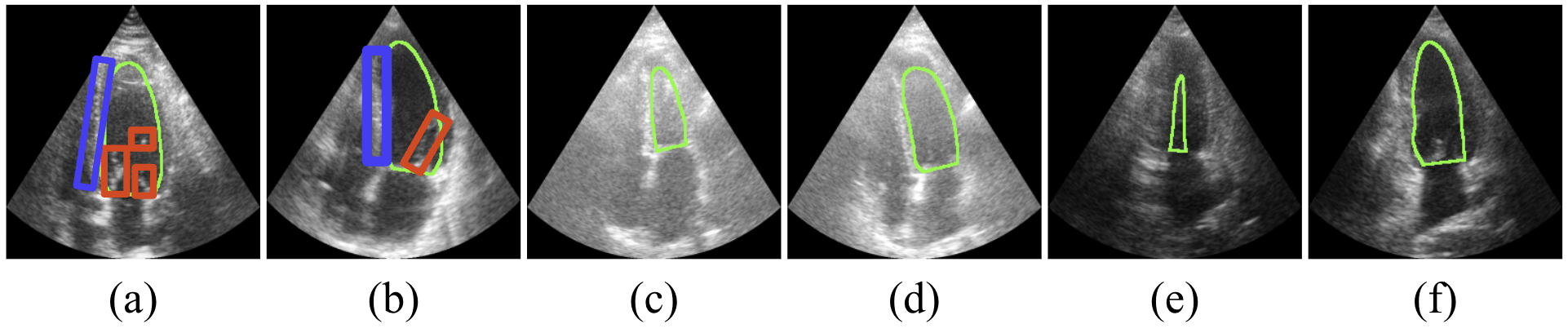}
    \caption{\textbf{Challenges} in echocardiography video segmentation: (a–b) red boxes of speckle noise and blue boxes indicate indistinct or blurred contours; (c–f) large shape and scale variations across the cardiac cycle.}
    \label{fig:cha}
\end{figure}

%% file: fig/Met/architecture.tex
\begin{figure*}[!t]
    \centering
    \includegraphics[width=0.95\textwidth]{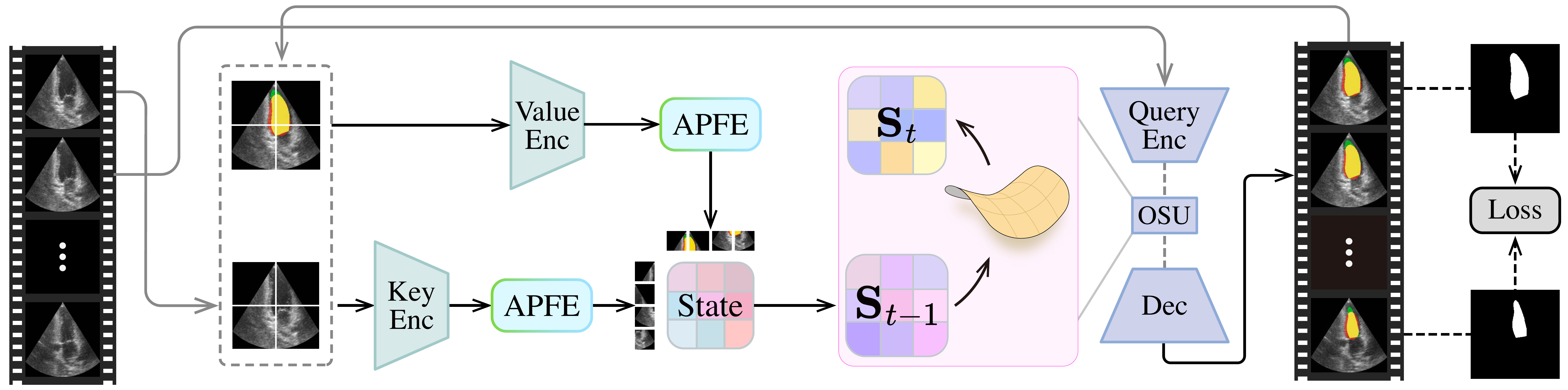}
\caption{Overview of the \mymodel architecture.
The Anatomical Prior-aware Feature Enhancement (APFE) encoder extracts contrast-decomposed features from echocardiography video frames, which are recurrently updated through the Orthogonalized State Update (OSU) mechanism to maintain stable and geometry-aware spatiotemporal representations.
The decoder reconstructs segmentation maps, and the entire model is optimized with loss.}
    \label{fig:met:architecture}
\end{figure*}

%% file: sec/2_0_related.tex
\section{Related Work}
    \label{sec:related}

\noindent
\textbf{Sequence modeling in echocardiography video segmentation.} 
Echocardiography video segmentation is significantly affected by factors such as speckle noise, probe motion, and rapid non-rigid cardiac deformations.
The standard temporal attention mechanism \cite{NIPS17_Attention,ICLR21_vit,TMI24_DSA} computes the output $o_t$ by calculating the correlation between the full history of key-value pairs $\{(\boldsymbol{k}_\tau, \boldsymbol{v}_\tau)\}_{\tau=1}^t$ and the current query $q_t$.
Although this mechanism can establish global temporal dependencies, its computational complexity grows quadratically with the sequence length.
\emph{Retrieval} paradigms based on a memory bank~\cite{iccv19_stm,neurips21_stm} have been adopted by methods such as XMem~\cite{eccv22_xmem}, XMem++~\cite{iccv25_xmempp}, Cutie~\cite{cvpr24_cutie}, and SAM 2 \cite{ICLR25_SAM2,CVPR25_DAM}.
Similarly, MemSAM~\cite{deng2024memsam} and SAMed-2~\cite{MICCAI25_SAMed2} maintain temporal consistency by filtering and storing reference frames in a memory bank, leveraging feature matching and retrieval.
However, limited by their discrete storage mechanisms, these methods struggle to fully utilize the continuous historical information of videos.
LRM \cite{katharopoulos_transformers_2020,ICML20_ttt,ICML21_fwp,ICLR25_gdn,ICLR26_ttt_LaCT} offer an alternative feature aggregation strategy: compressing the complete history into a fixed-size hidden state matrix $\mathbf{S}_t$, thereby achieving inference with constant time complexity.
EchoVim~\cite{mm25_echovim}, based on Vision Mamba \cite{CLM24_mamba,ICML24_mamba2,eccv24_videomamba}, introduces this type of architecture to improve computational efficiency.
However, its temporal modeling still relies on bidirectional inference dependent on specific frames and Dynamic Inter-frame Temporal Attention between adjacent frames, which essentially remains sparse modeling based on specific physiological priors.
Works targeting long-sequence video modeling (such as LiVOS~\cite{Liu_2025_CVPR} and GDKVM \cite{ICLR25_gdn,ICCV25_gdkvm}) utilize a linear key-value association mechanism to compress temporal history into a time-dependent hidden state $\mathbf{S}_t$.
However, these methods operate in unconstrained Euclidean space, where scalar gating mechanisms act as indiscriminate weight decay on the singular values of $\mathbf{S}_t$.
Over long sequences, this inevitably collapses the state matrix into a low-rank approximation, destroying fine-grained spatial details critical for valvular motion analysis and reducing the effective memory capacity.
Addressing this rank collapse requires fundamentally abandoning the Euclidean formulation in favor of geometric constraints that preserve the spectral structure of memory states.

\noindent
\textbf{Geometric constraints in recurrent optimization.}
Viewing sequence modeling through an optimization lens, methods like Test-Time Training \cite{ICML20_ttt,ICML25_ttt,arxiv25_ttregression} formulate hidden state updates as gradient descent steps on a surrogate objective.
However, because these updates operate purely in unconstrained Euclidean space, the state remains vulnerable to singular value degradation and rank collapse over long sequences.
In parallel, orthogonal constraints have shown benefits in parameter optimization.
Muon \cite{post24_muon,arxiv25_kimimuon,ICLR26_ttt_LaCT} applies Newton-Schulz iterations \cite{SIAM06_Schur,SIAM06_Newton} to orthogonalize the gradient or momentum rather than the weights themselves, finding the steepest descent direction under spectral norm constraints.
While effective for parameter updates, such constraints address the optimization dynamics rather than the inference-time state evolution.
Orthogonal and unitary constraints on state transitions have long been recognized as mechanisms to prevent rank collapse in traditional RNNs \cite{PUR08_Absil,ICML16_urnn,NIPS16_fullurnn}.
Extending this geometric rigor to the matrix-valued states of modern linear associative memory, however, has been hindered by the prohibitive computational cost of exact Singular Value Decomposition (SVD) for enforcing orthogonality.

%% file: sec/3_0_method.tex
\section{Method}
    \label{sec:method}

\noindent
We propose \mymodel, a sequence architecture designed to tackle the challenges of complex spatiotemporal information and ultrasound cardiac noise in echocardiography video segmentation (see~\cref{fig:met:architecture}).

Given a 2D ultrasound video sequence $\boldsymbol{I}_{1:t} \in \mathbb{R}^{tHW \times C_d}$, where $\boldsymbol{I}_{1:t}$ represents the sequence of ultrasound frames over time, the goal of our model is to predict the accurate LV segmentation masks for the entire sequence.
The model uses ResNet-50~\cite{He_CVPR16_resnet} as the visual backbone.
It employs APFE for contrast decomposition, processing the input frame $\mathbf{I}_t$ into a structured key representation $\mathbf{Key}_t$.
The model then constructs a fixed-size state representation $\mathbf{S}_t \in \mathbb{R}^{C_v \times C_k}$ using $\mathbf{Key}_t$ and its corresponding $\mathbf{Value}_t$.
During the linear temporal updates of $\mathbf{S}_t$, the model constrains the state evolution to the Stiefel manifold via an orthogonalized update mechanism, utilizing Newton-Schulz iteration for efficient computation.
Stability is further regularized through weight decay.
In the prediction phase, the features of the target frames (e.g., ED and ES frames) serve as $\mathbf{Query}_t$, interacting with the maintained state $\mathbf{S}_t$ to decode the final predictions.
Crucially, unlike semi-supervised video object segmentation methods that require a first-frame reference mask, our inference process operates in a fully automatic manner without relying on any manual prompt guidance, which is highly consistent with real-world clinical workflows.
During the training phase, the model adopts the evaluation setting from \cite{deng2024memsam,ICCV25_gdkvm} to simulate real-world clinical scenarios, optimizing by comparing predictions with ground truth annotations.

%% file: sec/3_1_OSU.tex
\subsection{Orthogonalized State Update}

\noindent
In sequence modeling for echocardiography video segmentation, preserving spatial details (\eg,~rapid valvular motion and subtle myocardial deformations) across long sequences is essential.
Standard linear recurrent models \cite{katharopoulos_transformers_2020,ICML20_ttt,ICML21_fwp,ICLR25_gdn,ICLR26_ttt_LaCT} encode historical information into a hidden state $\mathbf{S}_t \in \mathbb{R}^{C_v \times C_k}$, where we assume $C_v \ge C_k$.

The gated recurrent update \cite{ICLR25_gdn,ICCV25_gdkvm,arxiv25_kimidelta} corresponds to a proximal gradient descent step in unconstrained Euclidean space~\cite{FTO16_Hazan}.
At each timestep $t$, the incoming data introduces a linearized surrogate objective $\ell_t(\mathbf{S}) = -\operatorname{Tr}(\mathbf{G}_t^\top \mathbf{S})$, where $\mathbf{G}_t = \beta_t (\mathbf{v}_t - \alpha_t \mathbf{S}_{t-1} \mathbf{k}_t) \mathbf{k}_t^\top$ denotes the gradient.
The unconstrained state evolution minimizes this surrogate, regularized by a Euclidean proximity term:
\begin{equation}
  \begin{aligned}
    \mathbf{S}_t^{\text{Euc}} &= \mathbf{S}_{t-1}\left(\alpha_t (\mathbf{I}_{C_k} - \beta_t \mathbf{k}_t \mathbf{k}_t^\top)\right) + \beta_t \mathbf{v}_t \mathbf{k}_t^\top \\
    &= \underset{\mathbf{S} \in \mathbb{R}^{C_v \times C_k}}{\arg\min} \left( \ell_t(\mathbf{S}) + \frac{1}{2} \|\mathbf{S} - \alpha_t \mathbf{S}_{t-1}\|_F^2 \right).
  \end{aligned}
\end{equation}
However, performing gradient descent directly in unconstrained Euclidean space lacks intrinsic geometric regularization.
The gating mechanism $\alpha_t$ acts as an isotropic shrinkage.
In echocardiography video segmentation, the combination of this unconstrained shrinkage and the successive rank-1 data updates ($\mathbf{k}_t \mathbf{k}_t^\top$) alters the spectrum of the hidden state matrix.
Specifically, dominant update directions are amplified while orthogonal components decay, leading to rank collapse \cite{ICML16_urnn,NIPS16_fullurnn}.

\noindent
\textbf{Orthogonalized state update on the Stiefel manifold.}
To address this rank collapse, we propose \emph{Orthogonalized State Update}, which constrains the state evolution to the Stiefel manifold via an orthogonalized update
\begin{equation}
    \label{eq:stiefel_manifold}
    \mathcal{V}_{C_v, C_k} = \{\mathbf{S} \in \mathbb{R}^{C_v \times C_k} : \mathbf{S}^\top \mathbf{S} = \mathbf{I}_{C_k}\}.
\end{equation}
\begin{equation}
  \mathbf{S}_t = \operatorname{Proj}_{\mathcal{V}}(\mathbf{S}_t^{\text{Euc}}) = \underset{\mathbf{S} \in \mathcal{V}_{C_v, C_k}}{\arg\min} \frac{1}{2} \|\mathbf{S} - \mathbf{S}_t^{\text{Euc}}\|_F^2.
\end{equation}
By enforcing strict orthogonality, the Frobenius norm of the state remains constant ($\|\mathbf{S}_t\|_F^2 = C_k$).
Consequently, the projection is equivalent to maximizing the trace inner product: $\arg\max_{\mathbf{S} \in \mathcal{V}} \operatorname{Tr}(\mathbf{S}^\top \mathbf{S}_t^{\text{Euc}})$, which identifies the closest orthogonal matrix to $\mathbf{S}_t^{\text{Euc}}$.

        \input{fig/osu/osu}

As illustrated in~\cref{fig:osu}, the standard Euclidean recurrence (left) allows the state to deviate from the manifold, whereas OSU explicitly projects $\mathbf{S}_t^{\text{Euc}}$ back onto the Stiefel manifold to preserve orthogonality.
\Cref{fig:landscape_2d_behavior} further visualizes this from an optimization perspective: unconstrained updates deviate from the manifold constraint, while OSU projects the iterates back onto the feasible region.

\noindent
\textbf{High-order Newton-Schulz iteration for efficient projection.}
The exact solution computes the orthogonal polar factor via SVD~\cite{SIAM86_Higham}, incurring $\mathcal{O}(C_v C_k^2)$ cost.
To reduce this computational cost, we employ a parameterized high-order Newton-Schulz iteration \cite{SIAM06_Schur,SIAM06_Newton}, which converges to the closest orthogonal matrix faster than standard quadratic schemes.
Newton-Schulz iterations converge only when the initial singular values are strictly bounded.
Because the Frobenius norm upper-bounds the spectral norm ($\|\cdot\|_2 \le \|\cdot\|_F$), scaling the intermediate unconstrained state by its Frobenius norm provides a sufficient bound to ensure all singular values fall strictly within this convergence domain:
\begin{equation}
  \mathbf{X}^{(0)} = \frac{\mathbf{S}_t^{\text{Euc}}}{\|\mathbf{S}_t^{\text{Euc}}\|_F + \epsilon},
\end{equation}
where $\epsilon$ is a small constant ensuring numerical stability.
The orthogonalization is then performed using a 5th-order polynomial expansion:
\begin{equation}
  \mathbf{X}^{(j+1)} = a\mathbf{X}^{(j)} + b\mathbf{X}^{(j)}{\mathbf{X}^{(j)}}^\top \mathbf{X}^{(j)} + c\mathbf{X}^{(j)}({\mathbf{X}^{(j)}}^\top \mathbf{X}^{(j)})^2.
\end{equation}
The polynomial coefficients $a$, $b$, and $c$ are configured to optimize the spectral mapping function, thereby maximizing the rate of convergence towards the Stiefel manifold.
In practice, this iteration achieves orthogonality in a fixed number of steps, avoiding the exact SVD computation.

This orthogonalization stabilizes the representation spectrum.
In our sequence modeling implementation, projecting onto the Stiefel manifold acts as an effective constant spectral norm constraint ($\|\mathbf{S}_t\|_2 = \gamma$, where $\gamma$ is a non-zero scaling factor applied to the projected state).
By preventing singular value decay and bounding the condition number, our method avoids the rank collapse typically observed in unconstrained recurrent updates.
This scaled isometry maintains norm stability during state transitions, supporting long-range dependency modeling for echocardiography video segmentation by preserving consistent feature representations across complete cardiac cycles.

%% file: fig/osu/osu.tex
\begin{figure}[!t]
    \centering
    \includegraphics[width=0.94\linewidth]{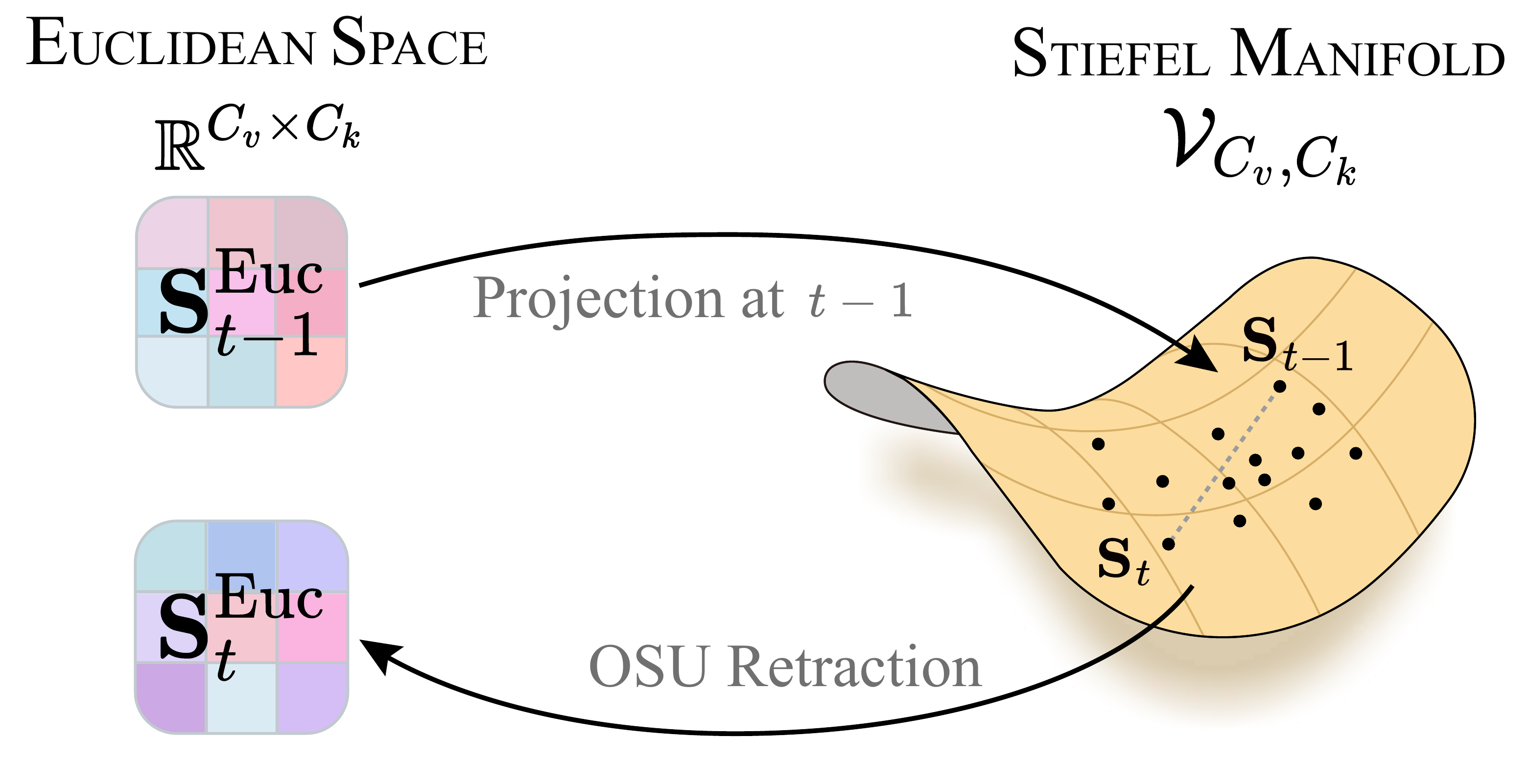}
    \caption{Illustration of OSU.
    Standard Euclidean recurrences suffer from rank collapse over long sequences (manifested as singular value decay and increasing condition number). OSU constrains state evolution on the Stiefel manifold via orthogonalized update. The unconstrained intermediate state $\mathbf{S}_t^{\text{Euc}}$ is projected back to the manifold as $\mathbf{S}_t$, ensuring stable temporal transitions.}
    \label{fig:osu}
\end{figure}

\begin{figure}[!t]
    \centering
    \includegraphics[width=1.0\linewidth]{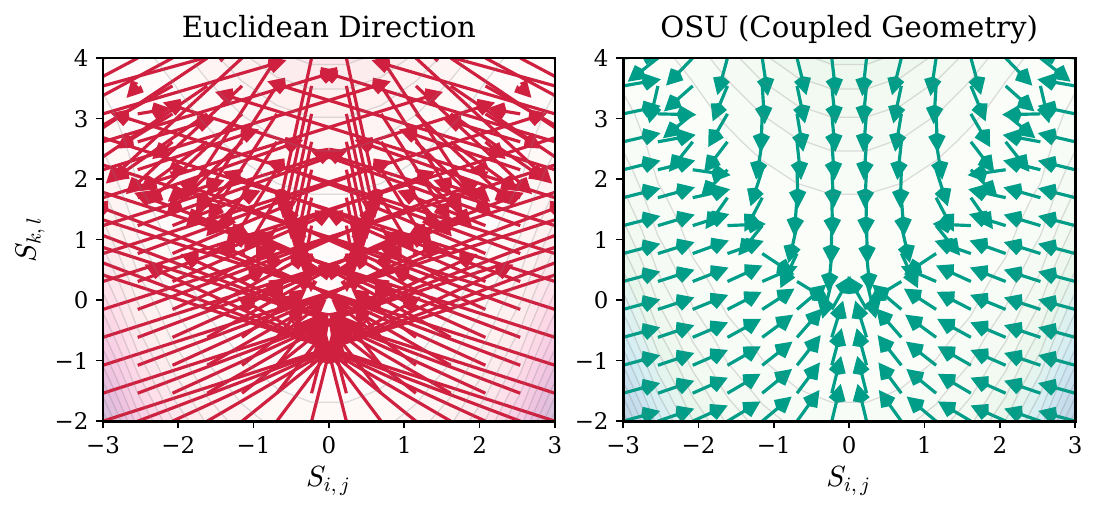}
    \caption{\textbf{Behavioral comparison of optimization vector fields.}
    Unconstrained Euclidean updates cause the state to drift away from orthogonal constraints (climbing valley walls). OSU generates a corrective flow that pulls iterates back onto the Stiefel manifold (valley floor).}
    \label{fig:landscape_2d_behavior}
\end{figure}

%% file: tab/cap.tex
\begin{table*}[!t]
    \centering
    \caption{Comparison of \mymodel with state-of-the-art methods for LV segmentation on the CAMUS~\cite{tmi19_camus} and EchoNet-Dynamic~\cite{nature20_echodynamic} datasets.
        Metrics include mean Dice coefficient (mDice) and mean 95\% Hausdorff distance (mHD95) for segmentation evaluation, as well as correlation coefficient (corr) and bias$\pm$std (\%) for $\text{LV}_{\text{EF}}$ estimation.}
    \label{tab:lv_segmentation_comparison}

    \small
    \setlength{\tabcolsep}{6pt}

    \begin{tabular}{l ccc r@{${}\pm{}$}l ccc r@{${}\pm{}$}l}
        \toprule

        \textbf{Method} & \multicolumn{5}{c}{\textbf{CAMUS}} & \multicolumn{5}{c}{\textbf{EchoNet-Dynamic}} \\
        \cmidrule(lr){2-6} \cmidrule(lr){7-11}

        & \textbf{mDice $\uparrow$} & \textbf{mHD95 $\downarrow$} & \textbf{corr $\uparrow$} & \multicolumn{2}{c}{\textbf{bias$\pm$std (\%)}} & \textbf{mDice $\uparrow$} & \textbf{mHD95 $\downarrow$} & \textbf{corr $\uparrow$} & \multicolumn{2}{c}{\textbf{bias$\pm$std (\%)}} \\
        \midrule

        PolaFormer~\cite{PolaFormer}~\textcolor{gray}{\scriptsize ICLR'25} &
        92.47 & 3.66 & 0.792 & -0.28 & 7.5 &
        91.51 & 3.89 & 0.678 & -0.01 & 8.1 \\

        Vision LSTM~\cite{ICLR2025_3b6eaef6}~\textcolor{gray}{\scriptsize ICLR'25} &
        92.71 & 3.47 & 0.745 & -2.36 & 7.6 &
        91.71 & 3.85 & 0.691 & -0.37 & 7.9 \\

        Cutie~\cite{cvpr24_cutie}~\textcolor{gray}{\scriptsize CVPR'24} &
        91.43 & 3.97 & 0.629 &  0.44 & 9.8 &
        90.44 & 4.05 & 0.600 &  0.40 & 9.6 \\

        LiVOS~\cite{Liu_2025_CVPR}~\textcolor{gray}{\scriptsize CVPR'25} &
        92.89 & 3.56 & 0.718 & -1.69 & 7.8 &
        91.92 & 3.82 & 0.702 & -0.04 & 7.5 \\
        \midrule

        SAMed-2~\cite{MICCAI25_SAMed2}~\textcolor{gray}{\scriptsize MICCAI'25} &
        93.55 & 3.29 & 0.815 & -2.16 & 6.8 &
        92.67 & 3.60 & 0.731 & -0.02 & 7.3 \\

        EchoVim~\cite{mm25_echovim}~\textcolor{gray}{\scriptsize ACM MM'25} &
        94.01 & 3.21 & 0.804 & -0.25 & 7.2 &
        93.22 & 3.47 & 0.803 & -0.62 & 6.2 \\

        Vivim~\cite{10973086}~\textcolor{gray}{\scriptsize TCSVT'25} &
        93.36 & 3.45 & 0.817 & -0.09 & 7.3 &
        92.49 & 3.70 & 0.758 &  1.43 & 7.0 \\

        GDKVM~\cite{ICCV25_gdkvm}~\textcolor{gray}{\scriptsize ICCV'25} &
        94.18 & 3.31 & 0.835 & -0.71 & 6.7 &
        93.33 & 3.55 & 0.810 & -0.08 & 6.5 \\
        \midrule

        \textbf{\mymodel} &
        \textbf{94.82} & \textbf{3.25} & \textbf{0.830} & \textbf{-0.44} & \textbf{7.0} &
        \textbf{93.90} & \textbf{3.42} & \textbf{0.816} & \textbf{-1.08} & \textbf{6.0} \\

        \bottomrule
    \end{tabular}
\end{table*}

%% file: sec/3_2_APFE.tex
\subsection{Anatomical Prior-aware Feature Enhancement}
\label{subsec:apfe}

Existing temporal models \cite{deng2024memsam,mm25_echovim,ICCV25_gdkvm} mainly rely on discrete memory or linear aggregation, treating ultrasound speckle and anatomical boundaries without distinction.
This approach lacks physical priors, leading to representation drift under rapid cardiac deformations and severe noise.
To address this issue, we propose the \emph{Anatomical Prior-aware Feature Enhancement} module (\cref{fig:apfe}).
Unlike standard visual encoders, APFE explicitly accounts for depth-dependent acoustic attenuation by decomposing the feature space into local acoustic fields and structural residuals, decoupling anatomical boundaries from stochastic noise and providing a stable basis for long-term propagation.

\noindent
\textbf{Intensity decoupling via local acoustic fields.}
Echocardiographic signals are corrupted by both stochastic speckle and severe depth-dependent acoustic attenuation \cite{8265568,WANG201925}.
The attenuation follows an approximately exponential decay with depth, which induces a spatially-varying acoustic bias field that confounds genuine tissue contrast.
Relying on a global threshold fails to account for this spatial heterogeneity.
Instead, we formalize the acoustic prior by decomposing the intermediate feature map $\mathbf{X}_t \in \mathbb{R}^{B\times C\times H\times W}$ into a low-frequency ambient acoustic field and high-frequency structural residuals.
Specifically, we estimate the spatially-varying acoustic bias field $\mathbf{M}_t \in \mathbb{R}^{B\times C\times H\times W}$ via large-kernel average pooling:
\begin{equation}
    \mathbf{M}_t = \mathrm{AvgPool}_{K \times K}(\mathbf{X}_t),
\end{equation}
where $K$ defines the receptive field of the acoustic environment.
By treating $\mathbf{M}_t$ as a local physical estimator of the acoustic bias field, we derive the polarity-aware components:
\begin{equation}
    \mathbf{X}_t^{+} = \mathrm{ReLU}(\mathbf{X}_t - \mathbf{M}_t), \quad \mathbf{X}_t^{-} = \mathrm{ReLU}(\mathbf{M}_t - \mathbf{X}_t).
\end{equation}
This dual-ReLU decomposition implicitly performs background-foreground decoupling: $\mathbf{X}_t^{+}$ isolates high-frequency structural edges (\eg, myocardium boundaries) by retaining only positive residuals above the local bias, while $\mathbf{X}_t^{-}$ captures low-response homogeneous regions (\eg, blood pools).
This formulation yields a lossless residual decomposition ($\mathbf{X}_t = \mathbf{X}_t^{+} - \mathbf{X}_t^{-} + \mathbf{M}_t$) that preserves all high-frequency details while filtering out low-frequency intensity fluctuations.

\input{fig/apfe/apfe}

\noindent
\textbf{Anatomical prior-aware fusion.}
To transform these decoupled components into high-level semantic features, we employ a dual-path strategy:
\begin{equation}
    \mathbf{H}_t^{+} = \phi^{+}(\mathbf{X}_t^{+}), \quad \mathbf{H}_t^{-} = \phi^{-}(\mathbf{X}_t^{-}),
\end{equation}
where $\phi^{+}$ and $\phi^{-}$ are unshared $3\times3$ Conv-BN-ReLU blocks.
This design enables dedicated pathways to separately handle structural geometry via $\phi^{+}$ and region-level semantics via $\phi^{-}$, respectively.
These specialized features are subsequently fused via an adaptive gating mechanism:
\begin{equation}
\label{eq:gating}
\begin{aligned}
    \lambda_t &= \sigma\!\left( \mathbf{W}_g \left[\mathbf{H}_t^{+};\, \mathbf{H}_t^{-}\right] \right), \\
    \mathbf{Z}_t &= \lambda_t \odot \mathbf{H}_t^{+} + (1 - \lambda_t) \odot \mathbf{H}_t^{-},
\end{aligned}
\end{equation}
where $\mathbf{W}_g$ denotes a $1\times1$ convolutional layer, $[\cdot ; \cdot]$ denotes channel-wise concatenation, and $\sigma$ is the sigmoid activation that produces a per-pixel attention map $\alpha_t \in \mathbb{R}^{B\times C\times H\times W}$.
By integrating APFE, the framework generates noise-resilient structural features $\mathbf{Z}_t$ that preserve fine anatomical boundaries decoupled from acoustic artifacts.
These refined features serve as stable key-value inputs for the subsequent sequence modeling process, enabling precise delineation of endocardial contours across varying imaging depths and cardiac phases.

%% file: fig/apfe/apfe.tex
\begin{figure}[t]
    \centering
        \includegraphics[width=1.0\linewidth]{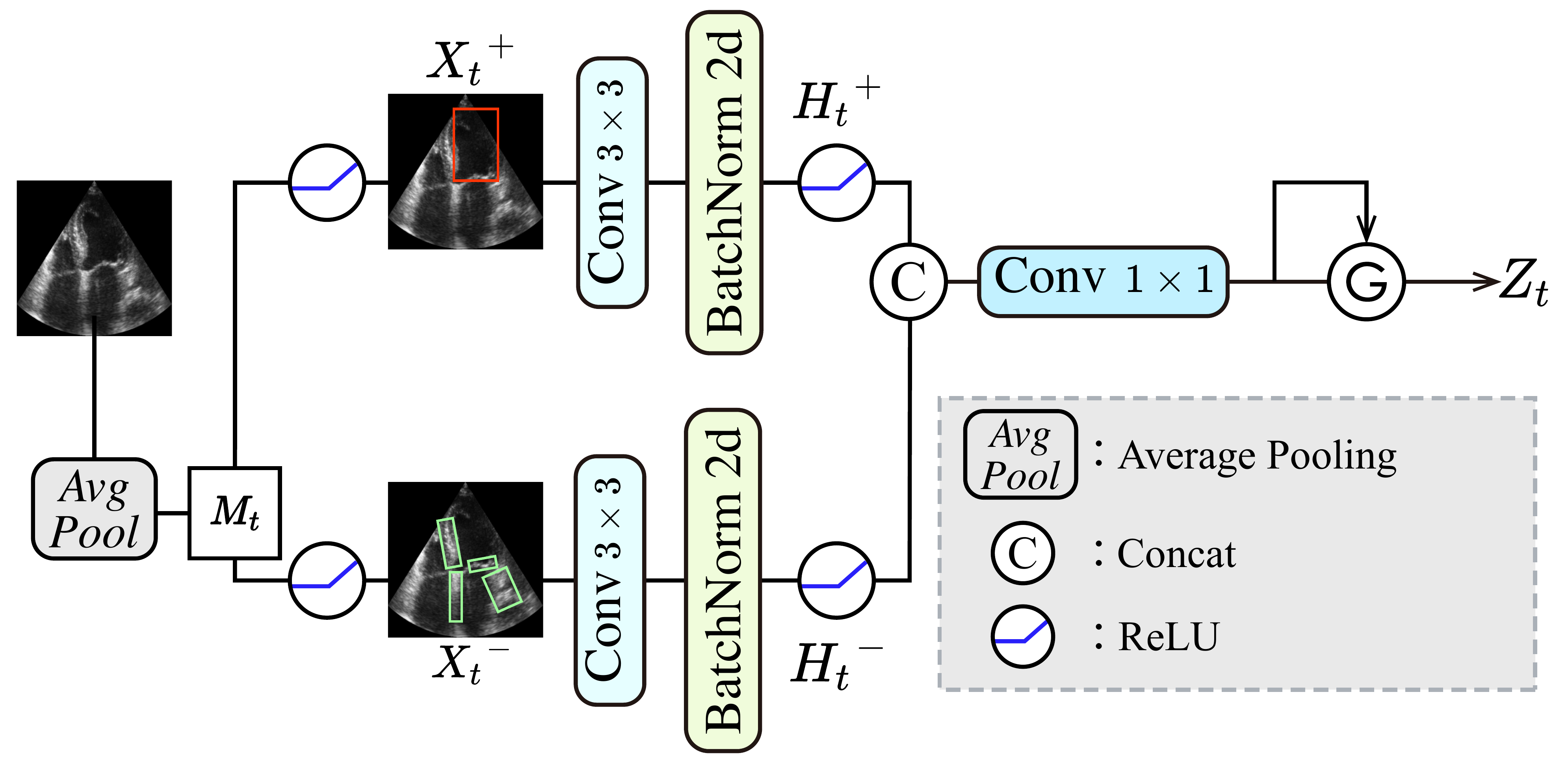}
    \caption{Illustration of the APFE module.
    APFE decomposes the input into high- and low-intensity components, processes them via dual convolutional branches, and fuses them through a gated mechanism to produce enhanced features $\mathbf{Z}_t$.}
    \label{fig:apfe}
\end{figure}

%% file: sec/4_0_exp.tex
\section{Experiments}
    \label{sec:experiments}

\subsection{Implementation Details}

\noindent
\textbf{Network architecture.}
We model temporal dynamics in ultrasound videos via a learnable state that evolves from frame-wise ResNet features refined by a lightweight APFE block. 
The state transition is stabilized through orthogonalization. 
Weight decay (0.02) is applied for regularization. 
During inference, frame features serve as $\mathbf{Query}_t$ to retrieve segmentation masks from the learned state $\mathbf{S}_t$, and training employs point-supervised cross-entropy and Dice loss following \cite{Liu_2025_CVPR,ICCV25_gdkvm} with AdamW~\cite{loshchilov2018decoupled_adamw} ($1\times10^{-4}$ learning rate, batch size $6$) and standard data augmentation on two RTX 2080 GPUs.

\noindent
\textbf{Datasets.}
We evaluate \mymodel on two public echocardiography video datasets: CAMUS~\cite{tmi19_camus} and EchoNet-Dynamic~\cite{nature20_echodynamic}. 
CAMUS contains 500 patient studies with apical four-chamber (A4C) and two-chamber (A2C) views.
Each frame from ED to ES is annotated with endocardial, epicardial, and atrial contours, offering dense supervision for assessing temporal consistency across the cardiac cycle. 
EchoNet-Dynamic includes 10,030 A4C videos with annotations only at ED and ES, supporting large-scale evaluation of model generalization under sparse labels. 
Combining these datasets enables comprehensive assessment under both dense and sparse annotation regimes, resembling real clinical settings where frame-wise labeling is costly. 
For preprocessing, CAMUS videos are resized to $256 \times 256$ with 15 frames, while EchoNet-Dynamic videos are resized to $128 \times 128$ with 10 frames.
We train \mymodel for 3000 iterations, sufficient for convergence on both datasets.

\noindent
\textbf{Metrics.}
We evaluate the model using standard segmentation and clinical performance metrics.
Geometric accuracy is quantified by the mean Dice coefficient (mDice) and the mean 95\% Hausdorff Distance (mHD95), averaged over the ED and ES phases.
For clinical relevance, we estimate the $\text{LV}_{\text{EF}}$ from the predicted masks using Simpson’s rule~\cite{LANG20151}.
The estimation accuracy is then evaluated using the Pearson correlation coefficient ($r$), mean bias, and standard deviation between the predicted and reference $\text{LV}_{\text{EF}}$ values.

%% file: tab/type.tex
\begin{table}[t]
    \centering
    \caption{Comparison of different models on CAMUS.
        \textbf{Paradigm} denotes the core temporal memory and update mechanism;
        \textbf{Para} is the number of parameters (M);
        \textbf{Mem} is the GPU memory usage during training (GB);
        and \textbf{Time} is the approximate training time (hours) to reach the best mDice.}
    \label{tab:type}
    \resizebox{\linewidth}{!}{
        \begin{tabular}{lcrrrr}
            \toprule
            \textbf{Method} & \textbf{Paradigm} & \textbf{Para} & \textbf{Mem} & \textbf{Time} \\
            \midrule
            PolaFormer & \textcolor{lime!70!black}{Linear} & 29.0 & 13.2 & 3.5 \\
            Vision LSTM & \textcolor{lime!70!black}{Linear} & 5.9 & 4.5 & 2.5 \\
            Cutie & \textcolor{orange!80!black}{Retrieval} & 39.1 & 8.1 & 6.0 \\
            LiVOS & \textcolor{green!70!black}{LKVA} & 35.1 & 7.4 & 5.0 \\
            SAMed-2 & \textcolor{orange!80!black}{Retrieval} & 110.1 & 27.5 & 7.0 \\
            EchoVim & \textcolor{lime!70!black}{SSM} & 42.0 & 34.9 & 9.0 \\
            Vivim & \textcolor{lime!70!black}{SSM} & 59.6 & 5.8 & 10.0 \\
            GDKVM & \textcolor{green!70!black}{LKVA} & 35.2 & 7.5 & 4.5 \\
            \midrule
            \textbf{\mymodel} & \textcolor{blue!70!black}{Manifold Proj.} & \textbf{38.3} & \textbf{7.6} & \textbf{3.0} \\
            \bottomrule
        \end{tabular}
    }
\end{table}

%% file: fig/Exp/vis_cap.tex
\begin{figure*}[!t]
    \centering
    \includegraphics[width=0.98\textwidth]{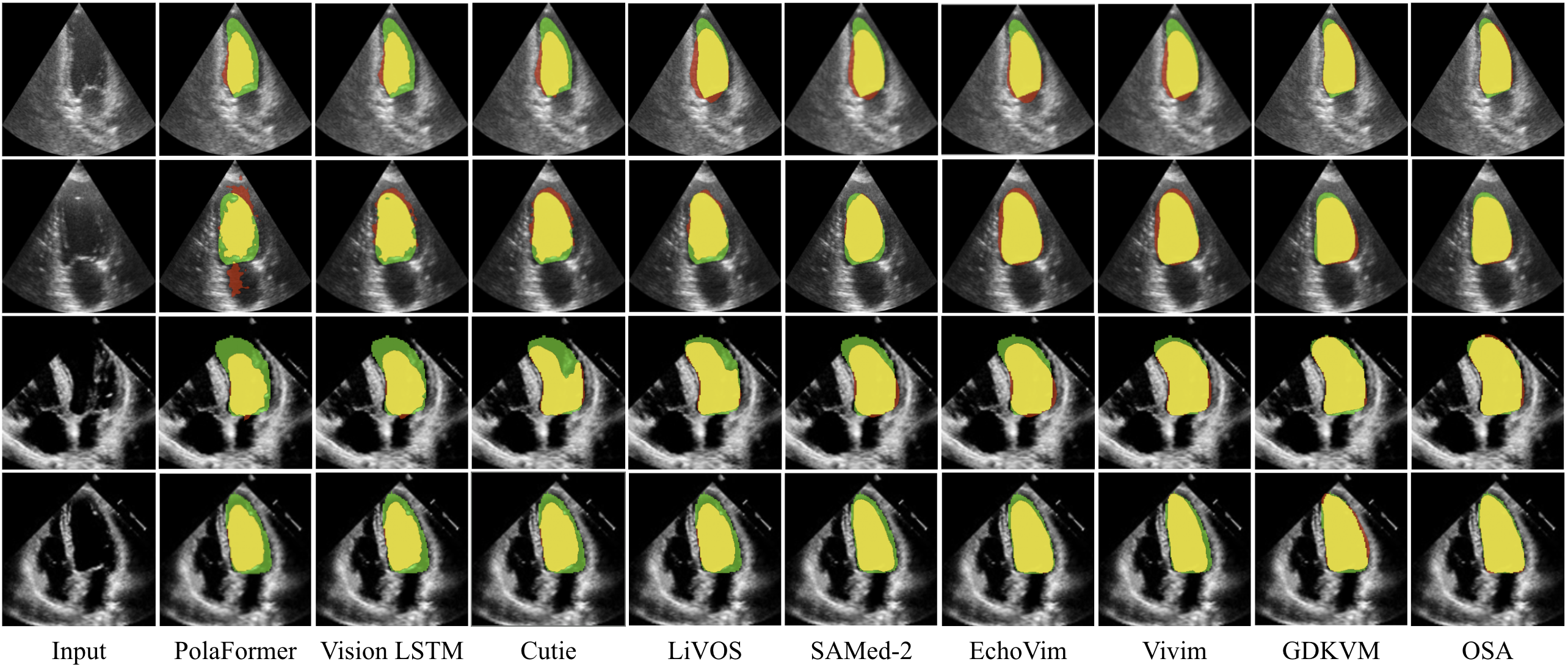}
    \caption{Visual comparison with state-of-the-art methods. The top two rows are from CAMUS, and the bottom two rows are from EchoNet-Dynamic. Green, red, and yellow regions represent the ground truth, prediction, and overlapping regions, respectively.}
    \label{fig:vis_cap}
\end{figure*}

%% file: sec/4_2_0_com_sota.tex
\subsection{Comparison with State-of-the-art Methods}
    \label{subsec:comparisons}

\noindent
We compared the performance of \mymodel with several state-of-the-art methods on the LV segmentation task.
As shown in~\cref{tab:type}, these methods employ different sequence modeling paradigms to address the challenges of LV segmentation,
including vision linear models (PolaFormer~\cite{PolaFormer}, Vision LSTM~\cite{ICLR2025_3b6eaef6}),
retrieval-based memory approaches (Cutie~\cite{cvpr24_cutie}, SAMed-2~\cite{MICCAI25_SAMed2}),
linear key-value association methods (LiVOS~\cite{Liu_2025_CVPR}, GDKVM~\cite{ICCV25_gdkvm}),
and state-space models (EchoVim~\cite{mm25_echovim}, Vivim~\cite{10973086}).
\mymodel adopts a manifold projection mechanism for stable temporal state evolution.

\noindent
\textbf{Quantitative results.}
\Cref{tab:lv_segmentation_comparison} shows that compared to traditional methods, \mymodel, through OSU and APFE, is able to capture deeper information in the temporal context, achieving the optimal balance between segmentation performance and computational cost.
In practical deployment scenarios, our model achieves \textbf{35~fps}, making real-time clinical segmentation possible.

\noindent
\textbf{Qualitative results.}
In~\cref{fig:vis_cap}, to ensure an objective evaluation, we select diverse, representative samples from CAMUS and EchoNet-Dynamic encompassing real-world clinical challenges like varying imaging qualities, acoustic shadowing, and ambiguous boundaries.
Across these conditions, \mymodel consistently yields more accurate ventricular contours and fewer artifacts than baseline methods.
The larger yellow overlap regions highlight its superior spatial agreement, demonstrating strong robustness and reliable generalization.

%% file: tab/abl.tex
\begin{table}[t]
    \centering
    \caption{Ablation study of the components on CAMUS.}
    \label{tab:abl}
    \begin{tabular}{lcccc}
        \toprule
        \textbf{Method}     & \textbf{mDice $\uparrow$} & \textbf{mHD95 $\downarrow$} & \textbf{Mem $\downarrow$} & \textbf{Time $\downarrow$} \\
        \midrule
        Baseline            & 92.94  & 3.56 & 7.5 & 4.5 \\
        w/o APFE            & 93.61  & 3.29 & 7.6 & 3.0 \\
        w/o OSU             & 94.12  & 3.21 & 7.5 & 4.7 \\
        \midrule
        Full                & 94.82  & 3.25 & 7.6 & 3.0 \\
        \bottomrule
    \end{tabular}
\end{table}

%% file: fig/abl/abl.tex
\begin{figure}[t]
    \centering
        \includegraphics[width=0.95\linewidth]{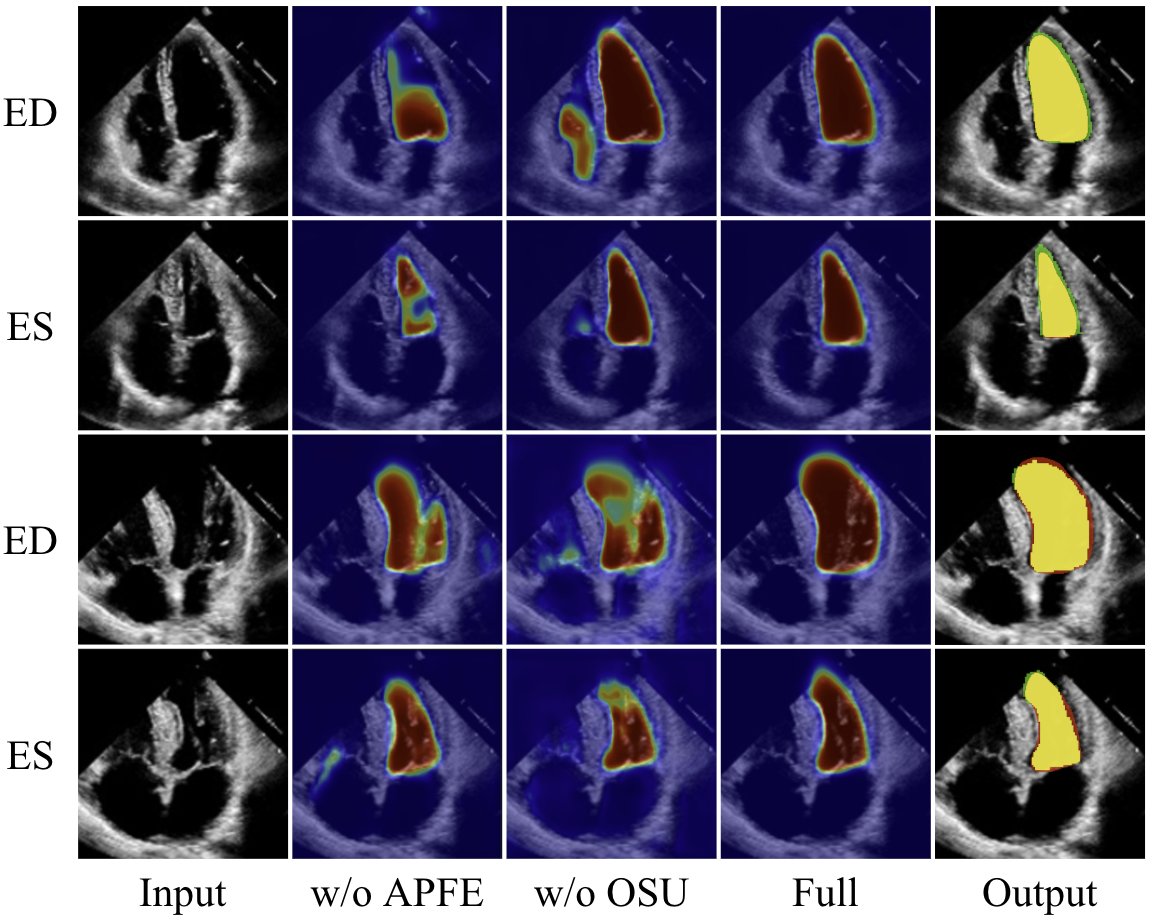}
    \caption{Heatmap ablation on EchoNet-Dynamic. Warmer colors indicate higher prediction confidence.}
    \label{fig:abl}
\end{figure}

%% file: sec/4_3_0_abl.tex
\subsection{Ablation Studies}
\label{subsec:abl}

\noindent
We conduct ablation studies on the network structure of \mymodel to validate the contribution of each component.
As detailed in~\cref{tab:abl}, removing both the APFE module and the OSU mechanism degrades our architecture into a naive linear key-value association model (Baseline).
Lacking geometric regularization and anatomical constraints, this baseline struggles to maintain temporal coherence and representation fidelity, resulting in noticeably limited segmentation accuracy.

\noindent
\textbf{Effect of APFE.}
\Cref{tab:abl,fig:abl} demonstrate that the APFE module consistently improves the segmentation performance of \mymodel.
As shown in~\cref{tab:abl}, the Full model achieves 94.82 mDice, a +1.88 improvement over the naive Baseline (92.94).
APFE contributes +0.67 (93.61 vs 92.94) while OSU contributes the remaining +1.21 (94.82 vs 93.61), with negligible memory overhead during training.
APFE injects structural priors into feature aggregation, enabling the model to capture cardiac morphology and preserve anatomical boundaries against ultrasound artifacts (\eg, speckle noise, attenuation).
Visualizations in~\cref{fig:abl} confirm that APFE produces sharper activation responses along the endocardial borders, highlighting its superior capability in modeling complex cardiac geometry.

\input{tab/abl_stab}

\noindent
\textbf{Effect of OSU.}
\Cref{tab:geo_stability,tab:num_stability} validate our theoretical claims regarding structural regularization.
Without manifold constraints, the Baseline model suffers from severe ill-conditioning and rank collapse: its minimum singular values vanish in 91.40\% of the steps (ColR), while the Mean Singular Value (MSV) uncontrollably drifts to 6.82 with high variance (SVVar = 1.47).
By incorporating OSU, the network achieves strictly conditioned temporal propagation.
The iterative gradient flow maps the state matrix to a scaled Stiefel manifold, forcing the Singular Value Variance (SVVar) to identically 0.00 and stabilizing the MSV at a constant 2.00.
This isotropic state provides a well-conditioned geometric foundation for sequence modeling~\cite{ICML16_urnn}.
Furthermore, the sharp decreases in Orthogonality Error (OrthE) and Temporal Drift provide direct empirical evidence that our approximate orthogonalization enforces consistent structural evolution.
By maintaining this strict isometry, OSU suppresses motion artifacts and preserves anatomical fidelity throughout the entire cardiac cycle.

%% file: tab/abl_stab.tex
\begin{table}[!t]
    \centering
    \setlength{\tabcolsep}{4.5pt}
    \caption{\textbf{Geometric stability.}
        MSV/SVVar measure spectral behaviour; OrthE quantifies orthogonality deviation;
        ColR (\%) is the percentage of steps with $\sigma_{\min}<10^{-3}$.}
    \label{tab:geo_stability}
    \begin{tabular}{
            l
            S[table-format=1.2]
            S[table-format=1.2]
            S[table-format=2.2]
            S[table-format=2.2]
        }
        \toprule
        Method & {MSV $\downarrow$} & {SVVar $\downarrow$} & {OrthE $\downarrow$} & {ColR $\downarrow$} \\
        \midrule
        Baseline      & 6.82 & 1.47 & 21.30 & 91.40 \\
        w/o Ortho.    & 2.91 & 0.63 & 15.80 & 88.10 \\
        w/o Spectral  & 3.77 & 1.12 &  9.74 & 41.60 \\
        \midrule
        \textbf{Full} & {\textbf{2.00}} & {\textbf{0.00}} & {\textbf{8.48}} & {\textbf{77.78}} \\
        \bottomrule
    \end{tabular}
\end{table}

\begin{table}[!t]
    \centering
    \setlength{\tabcolsep}{4.5pt}
    \caption{\textbf{Numerical stability.}
        GradVar measures gradient variance, UpdateNorm Var quantifies update-scale fluctuations,
        MeanSV reflects state magnitude, and Drift measures long-horizon deviation.}
    \label{tab:num_stability}
    \begin{tabular}{l*{4}{c}}
        \toprule
        Method & GVar $\downarrow$ & UpdVar $\downarrow$ & MSV $\downarrow$ & Drift $\downarrow$ \\
        \midrule
        Baseline      & 4.7e$^{-3}$ & 1.82 & 6.82 & 91.4 \\
        w/o WD/RMS    & 1.9e$^{-3}$ & 0.97 & 2.91 & 57.0 \\
        w/o Spectral  & 8.3e$^{-4}$ & 0.41 & 3.77 & 42.5 \\
        \midrule
        \textbf{Full} & \textbf{2.5e$^{-4}$} & \textbf{0.12} & \textbf{2.00} & \textbf{36.0} \\
        \bottomrule
    \end{tabular}
\end{table}

%% file: sec/5_limitations.tex
\subsection{Discussions and Limitations}
    \label{subsec:discussion}

\noindent
While \mymodel improves temporal consistency, it has several practical caveats.
Under the LRM architecture, the linear attention enhances the representational capacity of the state matrix through key-value joint modeling~\cite{katharopoulos_transformers_2020}; gated delta net \cite{JPA88_sinnm,EL89_nncudr,nc92_schmidhuber,ICLR25_gdn,ICCV25_gdkvm,arxiv25_kimidelta} optimizes the update formula; and we push this forward by shifting the state transition process onto a manifold.
However, despite improving temporal consistency, our training regime remains constrained by fixed-length sequences.
Since real-world medical videos are continuous with varied durations, maintaining stable memory propagation in a fully online setting remains challenging.
We will explore further optimization directions within the LRM framework in the future.
The method may produce failure cases in some situations (\cref{fig:fail}), which could be related to factors such as initialization.
Furthermore, evaluation on datasets with consistent imaging protocols may lack robustness against real-world domain shifts caused by varying ultrasound equipment, probe orientations, or image qualities.
Future work will focus on improving cross-domain generalization and adapting to continuous, unseen clinical video distributions.

%% file: fig/fail/fail.tex
\begin{figure}[t]
    \centering
        \includegraphics[width=0.75\linewidth]{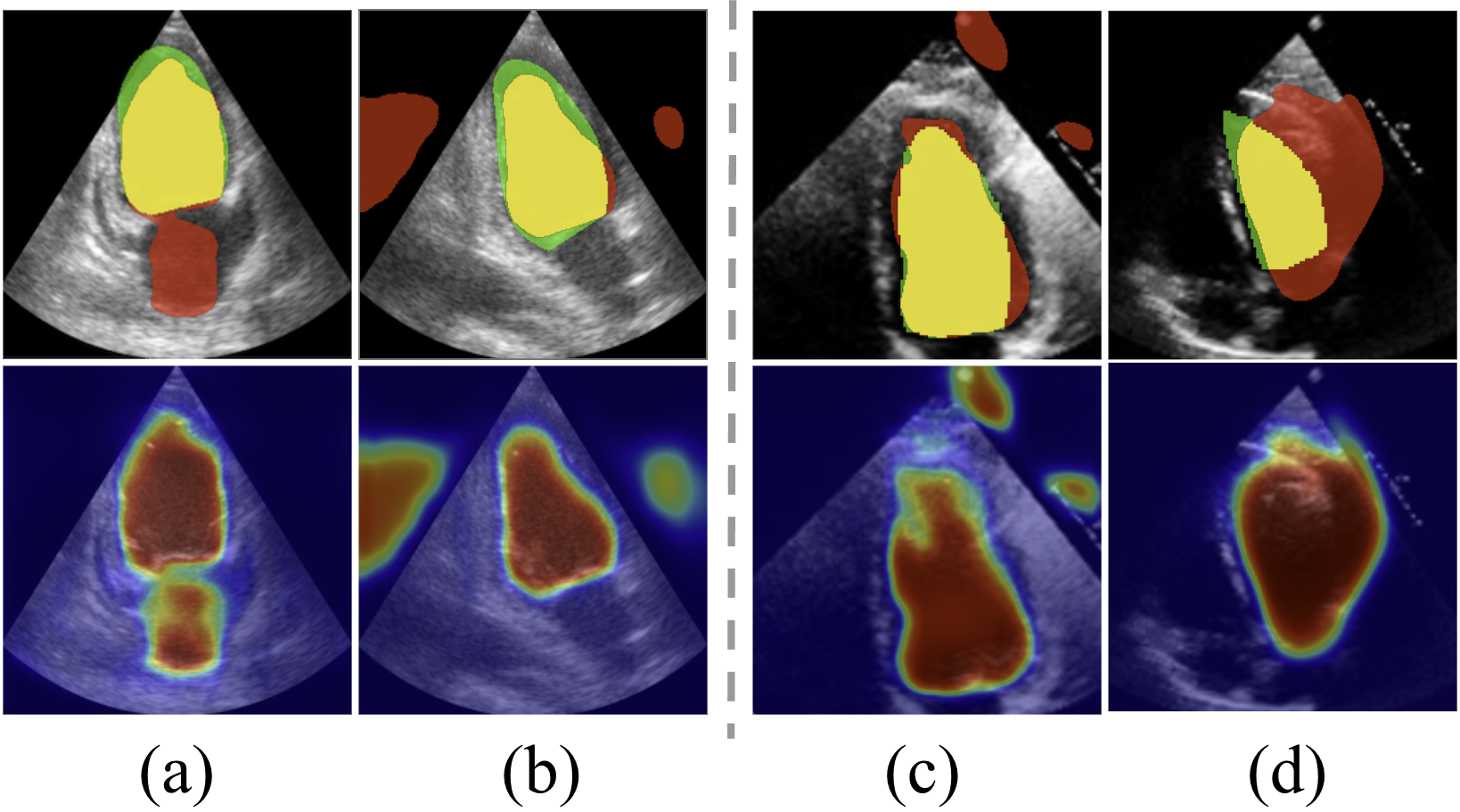}
    \caption{Failure cases on CAMUS (a-b) and EchoNet-Dynamic (c-d).}
    \label{fig:fail}
\end{figure}

%% file: sec/6_conclusion.tex
\section{Conclusion}

\noindent
In this paper, we proposed \mymodel, a novel framework designed to address the unique challenges of echocardiography video segmentation, specifically severe speckle noise and rank collapse in linear recurrent models. 
The core contribution includes two components: OSU, which constrains the memory evolution to the Stiefel manifold via an orthogonalized update to prevent rank collapse and maintain stable temporal transitions; and APFE, which explicitly separates anatomical structures from speckle noise through a physics-driven process, providing noise-resilient structural cues for temporal tracking.
Extensive experiments on CAMUS and EchoNet-Dynamic demonstrate that \mymodel achieves state-of-the-art segmentation accuracy and temporal stability.
Additionally, by maintaining real-time inference efficiency, our approach supports ejection fraction estimation and cardiac function assessment for practical clinical deployment.

%% file: sec/ackn.tex
\section*{Acknowledgments}
This work was supported partly by National Natural Science Foundation of China (No.~62273241), Natural Science Foundation of Guangdong Province, China (No. 2024A1515011946), the Shenzhen Research Foundation for Basic Research, China (No. JCYJ20250604181940054), and the grant under Hong Kong RGC Collaborative Research Fund (project no C5055-24G).

%% file: xsec/X_suppl.tex
\clearpage
\setcounter{page}{1}
\def\theHpage{Supplement.\arabic{page}}
\renewcommand{\thepage}{A.\arabic{page}}
\maketitlesupplementary

\appendix

\setcounter{section}{0}
\renewcommand{\thesection}{\Alph{section}}
\renewcommand{\theHsection}{\Alph{section}}

\crefname{section}{Appendix}{Appendices}
\Crefname{section}{Appendix}{Appendices}
\crefname{subsection}{Appendix}{Appendices}
\Crefname{subsection}{Appendix}{Appendices}
\setcounter{table}{0}
\setcounter{equation}{0}
\setcounter{figure}{0}
\renewcommand{\thetable}{A.\arabic{table}}
\renewcommand{\theequation}{A.\arabic{equation}}
\renewcommand{\thefigure}{A.\arabic{figure}}
\renewcommand{\theHtable}{A.\arabic{table}}
\renewcommand{\theHequation}{A.\arabic{equation}}
\renewcommand{\theHfigure}{A.\arabic{figure}}

\addcontentsline{toc}{part}{Appendix} 
\startcontents[supp]
\printcontents[supp]{l}{1}{
    \subsection*{Contents}
    \setcounter{tocdepth}{2}
}

% ------
\input{xsec/vis_com_osu}
                \input{fig/landscape/landscape}
\input{xsec/problem_setting}
\input{xsec/exp_con}
                \input{fig/train_step/train_step}
\input{xsec/efe}
\input{xsec/temporal_consistency}
% \input{xsec/rw_video_0}

%% file: xsec/vis_com_osu.tex
\section{Visualizing Optimization Dynamics}
\label{sec:vis_opt}

We analyze the geometric properties of the proposed Orthogonalized State Update (OSU) by comparing its optimization trajectories with standard Euclidean gradients.

\paragraph{Manifold-constrained optimization.}
As illustrated in~\cref{fig:landscape_3d_geo}, Euclidean updates operate in unconstrained space, treating the state matrix as a flattened vector.
This approach risks violating the orthogonality constraints of the Stiefel manifold.
In contrast, OSU employs an orthogonalized update via Newton-Schulz iteration, first performing a standard gradient step in ambient space then projecting back to the manifold, ensuring strict orthogonality of the state.
By enforcing strict orthogonality of the state $\mathbf{S}$, OSU ensures numerical stability and prevents rank collapse over long sequences.

\paragraph{Convergence stability.}
The vector fields in~\cref{fig:landscape_2d_behavior} demonstrate that Euclidean gradients exhibit high-frequency oscillations due to a lack of manifold curvature awareness.
In video segmentation tasks, such instabilities result in temporal inconsistency across frames.
Conversely, OSU functions as an implicit preconditioner, aligning the gradient flow with the optimization landscape.
This results in smoother convergence trajectories, which are essential for modeling continuous physiological dynamics, such as the cardiac cycle, without introducing high-frequency noise into the latent representation.

\paragraph{Feature decorrelation and consistency.}
The orthogonality constraints imposed by OSU promote semantic decorrelation within the latent state.
By enforcing $\mathbf{S}^\top \mathbf{S} = \mathbf{I}$, OSU reduces feature redundancy and prevents mode collapse.
This structural prior enables distinct state dimensions to represent independent dynamics, such as separating low-frequency ventricular contractions from high-frequency valve mechanics.
Unlike the Euclidean regime where feature interference leads to temporal instability, OSU ensures that diverse motion patterns evolve along orthogonal trajectories, thereby improving the temporal consistency of the resulting segmentation masks.

%% file: fig/landscape/landscape.tex
\begin{figure}[!t]
    \centering
    \includegraphics[width=0.98\linewidth]{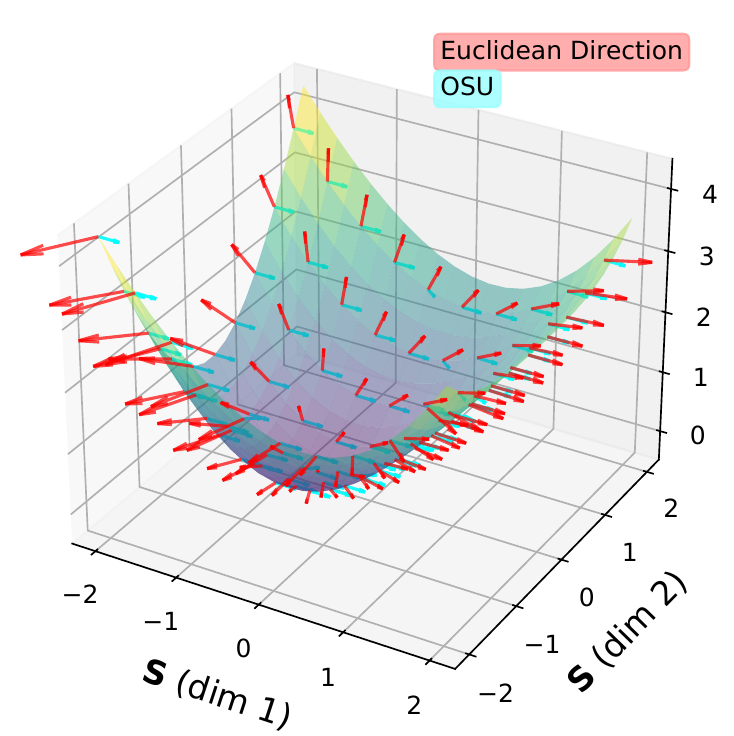}
    \caption{\textbf{3D geometric intuition} of update directions on the loss landscape.
    The \hlred{Euclidean Direction} updates the state element-wise along the steepest descent, often ignoring local curvature.
    \hlcyan{OSU} evolves the state by projecting gradients via the matrix sign function, thereby maintaining the consistency of the underlying subspace structure.
    The $x$ and $y$ axes represent the two dimensions of the state $\mathbf{S}$.
    The movement toward the lower regions of the surface represents the direction of loss optimization.}
    \label{fig:landscape_3d_geo}
\end{figure}

%% file: xsec/problem_setting.tex
\section{Problem Setting}

\noindent 
Exhaustive frame-by-frame annotation of echocardiography videos is labor-intensive in clinical settings.
Consequently, this work addresses semantic segmentation under a sparse temporal supervision setting~\cite{ICCV25_gdkvm}.

Let the input echocardiography video sequence of length $T$ be denoted as
\begin{equation*}
    V = \{x_1, x_2, \dots, x_T\}.
\end{equation*}
where $x_t$ represents the image frame at time step $t$. 

Under the target sparse supervision setting, ground truth segmentation masks are provided exclusively for the first ($t=1$) and the last ($t=T$) frames, which typically correspond to the end-diastole and end-systole phases.
Thus, the index set of annotated frames is strictly defined as
\begin{equation*}
    S = \{1, T\}.
\end{equation*}

The available ground truth segmentation masks during the training phase are defined as
\begin{equation*}
    Y_S = \{y_1, y_T\}.
\end{equation*}
\noindent
This formulation indicates that labels for frames $1 < t < T$ are unavailable during training.

The objective is to train a segmentation model $f_\theta$ parameterized by $\theta$.
The model processes the entire video sequence $V$ to generate dense segmentation predictions for all frames
\begin{equation*}
    \hat{Y} = \{\hat{y}_1, \hat{y}_2, \dots, \hat{y}_T\} = f_\theta(V).
\end{equation*}

Given the sparse annotations, the objective function is computed exclusively on the first and last frames.
The loss $\mathcal{L}$ is formulated as
\begin{equation*}
    \mathcal{L}(\theta) = \ell(\hat{y}_1, y_1) + \ell(\hat{y}_T, y_T),
\end{equation*}
where $\ell$ denotes a standard segmentation loss function.
Optimizing this objective requires the model to utilize temporal context to generate segmentations for the unannotated frames.
During the inference phase, the model operates in a fully automated manner, generating predictions for the entire sequence without requiring any additional prompts or manual interventions, simulating a real-world clinical workflow.

%% file: xsec/exp_con.tex
\section{Experiments Continued}
    \label{sec:exp_con}

\noindent 
\textbf{Segmentation evaluation.} For a fair comparison, the segmentation metrics (mDice, mHD95) reported in~\cref{tab:lv_segmentation_comparison} are evaluated exclusively on the annotated ED and ES frames, consistent with the sparse supervision protocol. 

\noindent
\textbf{Data and calibration.} All results for the medical metrics in~\cref{tab:lv_segmentation_comparison} are derived from the official CAMUS test subset.
Notably, we report the raw correlation, bias, and standard deviation without applying any bias-correction or post-calibration techniques to reflect the model's direct predictive performance.

\noindent
\textbf{Training dynamics and convergence.}
As illustrated in~\cref{fig:training_progression}, \mymodel exhibits a clear transition in both segmentation accuracy and feature localization.
In the initial phase (iterations 10--30), the segmentation masks are characterized by coarse boundaries, while the corresponding feature activation maps~(\cref{fig:training_iter10,fig:training_iter20,fig:training_iter30}, bottom rows) show diffused activations across the spatial domain.
By iteration 1250 (epoch 19), the model produces spatially precise segmentations~(\cref{fig:training_iter1250}), with feature activations concentrated strictly on the target anatomical structures.
This progression indicates that the training protocol facilitates the transition from global spatial search to fine-grained structural delineation without performance degradation.

\noindent
\textbf{Temporal consistency analysis.}
The visualization across the frame sequences in~\cref{fig:training_progression} demonstrates the capacity of the model to maintain temporal stability.
In the early stages, predictions are temporally inconsistent, with visible fluctuations in mask geometry across frames.
However, the converged model~(\cref{fig:training_iter1250}) preserves structural identity and boundary continuity throughout the cardiac cycle.
This stability is attributed to the following mechanisms.

\noindent
\textbf{Spatiotemporal feature aggregation.} The integration of contextual information from the temporal sequence facilitates the reduction of frame-wise prediction noise by utilizing the temporal evolution of anatomical structures.

\noindent
\textbf{Gradient propagation.} Backpropagating the loss through the temporal network enforces a consistency constraint on intermediate frames, ensuring that non-annotated frames align with the patterns learned from supervised frames.

\noindent
\textbf{Sequential context integration.} The refinement of internal representations as the sequence progresses allows the model to utilize cumulative temporal evidence for more accurate boundary localization, effectively correcting initial segmentation noise.

As shown in~\cref{fig:training_progression}, the segmentation quality of intermediate frames without direct supervision remains consistent with the supervised frames.
This indicates that the architecture effectively enforces temporal consistency and preserves structural identity throughout the sequence.

%% file: fig/train_step/train_step.tex
\begin{figure*}[!t]
    \centering
    % Row 1
    \begin{subfigure}[b]{1\textwidth}
        \centering
        \includegraphics[width=\textwidth]{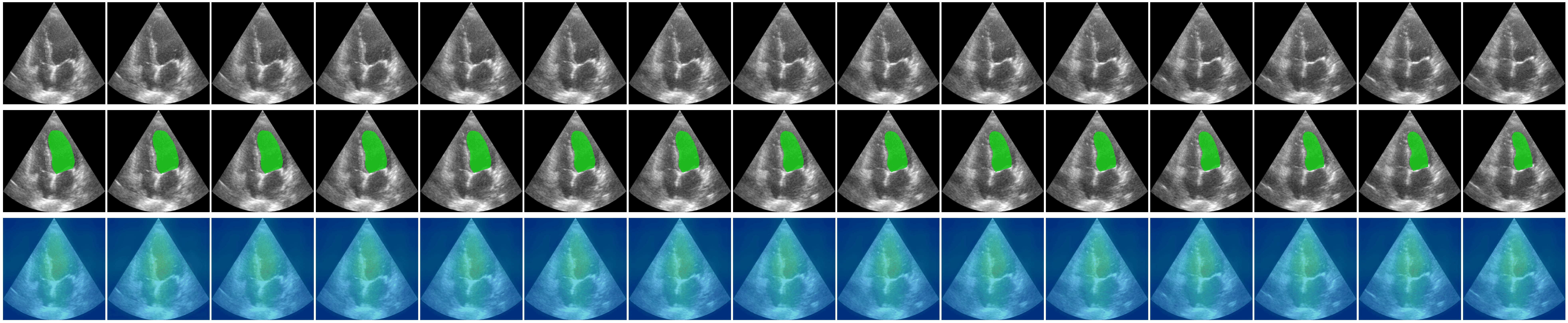}
        \caption{Iteration 10, Epoch 1}
        \label{fig:training_iter10}
    \end{subfigure}
    % Row 2
    \begin{subfigure}[b]{1\textwidth}
        \centering
        \includegraphics[width=\textwidth]{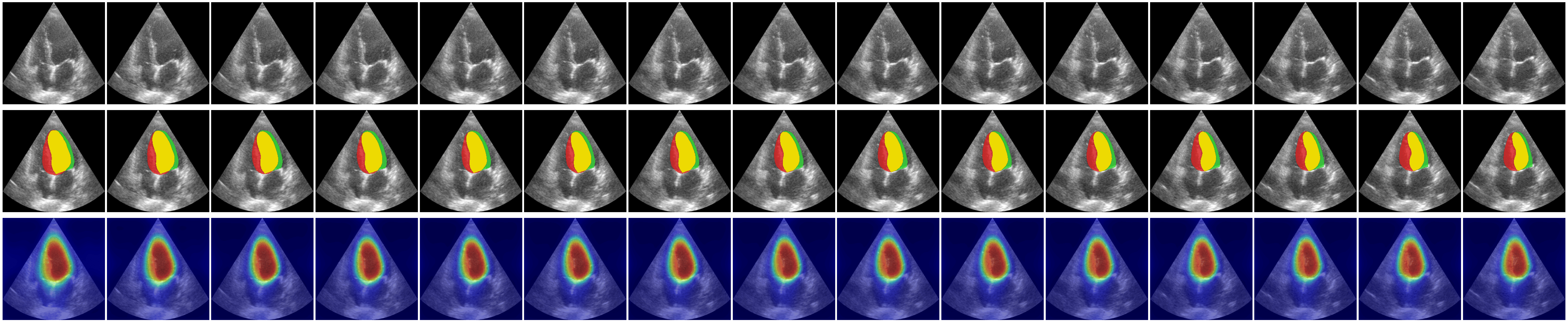}
        \caption{Iteration 20, Epoch 1}
        \label{fig:training_iter20}
    \end{subfigure}
    % Row 3
    \begin{subfigure}[b]{1\textwidth}
        \centering
        \includegraphics[width=\textwidth]{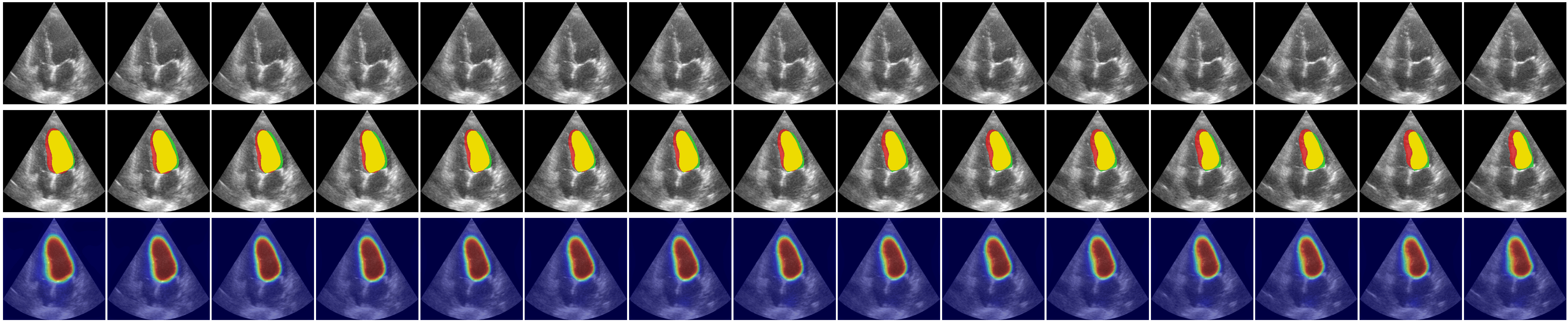}
        \caption{Iteration 30, Epoch 1}
        \label{fig:training_iter30}
    \end{subfigure}
    % Row 4
    \begin{subfigure}[b]{1\textwidth}
        \centering
        \includegraphics[width=\textwidth]{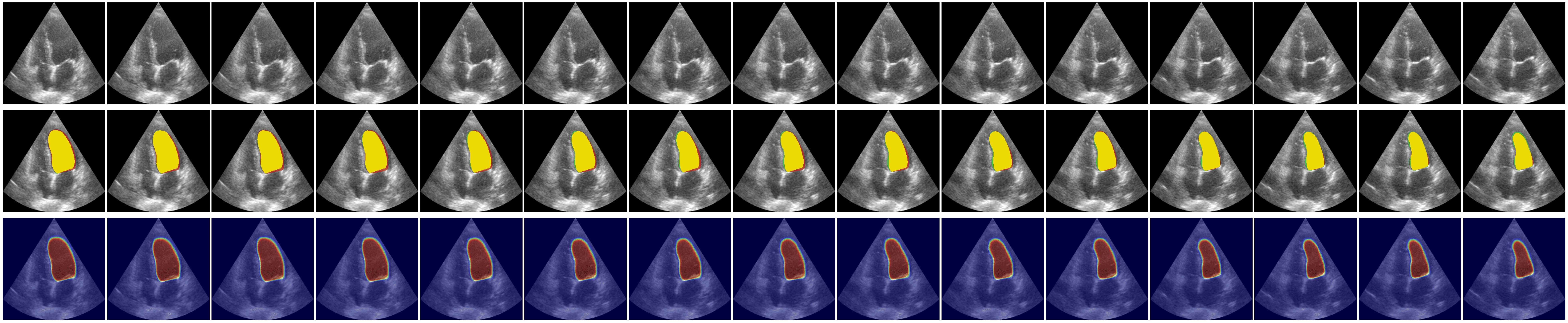}
        \caption{Iteration 1250, Epoch 19}
        \label{fig:training_iter1250}
    \end{subfigure}
    \caption{\textbf{Spatiotemporal training dynamics visualization.}
    Qualitative results on CAMUS patient 0225 (4-chamber view) at four training stages. For each stage (a-d), the top row displays input frames with segmentation masks, and the bottom row shows the corresponding feature activation maps. The progression from (a) to (d) illustrates the transition from diffused spatial activations and coarse masks to localized feature representations and temporally consistent segmentations throughout the cardiac cycle.}
    \label{fig:training_progression}
\end{figure*}

%% file: xsec/efe.tex
\section{Clinical Applications}
      \label{sec:clinical_application}

\noindent 
The Left Ventricular Ejection Fraction is a critical metric for cardiac function assessment.
Traditionally, it requires accurate measurement of End-Diastolic Volume ($V_{\text{ED}}$) at peak filling and End-Systolic Volume ($V_{\text{ES}}$) at peak contraction.
The standard volumetric $\text{LV}_{\text{EF}}$ formula is defined as:
\begin{equation*}
   \text{LV}_{\text{EF}} = \frac{V_{\text{ED}} - V_{\text{ES}}}{V_{\text{ED}}} \times 100\%.
   \label{eq:lvef}
\end{equation*}

While the Biplane Method of Disks (modified Simpson's rule) is the clinical gold standard for computing these volumes, it requires consistent dual-view data.
To ensure compatibility with single-view datasets (\eg,~EchoNet-Dynamic) and to directly leverage our automated segmentation framework, we employ a surrogate EF based on 2D area changes:
\begin{equation*}
   EF_{area} = \frac{A_{ED} - A_{ES}}{A_{ED}}.
   \label{eq:ef_area}
\end{equation*}
where $A_{ED}$ and $A_{ES}$ denote the segmented Left Ventricular areas at end-diastole and end-systole, respectively.
This surrogate metric is calculated independently for apical 4-chamber and 2-chamber views without biplane fusion.

By utilizing this area-based approach, our method provides a direct and robust pathway from automated single-view echocardiographic segmentation to actionable diagnostic metrics, enabling efficient and clinically relevant cardiac assessment.

%% file: xsec/temporal_consistency.tex
\section{Temporal Consistency in Video Segmentation}
\label{sec:temporal_consistency}

\noindent 
Temporal consistency in video segmentation is defined as the stability of predicted masks across consecutive frames.
While optical flow-based warping is a standard approach for quantifying such stability, its efficacy is constrained in echocardiography due to significant speckle noise, which degrades motion estimation accuracy.
To address this, the temporal matching error $\mathcal{E}_{tme}$ is employed, defined as the mean absolute difference between the temporal Dice evolution of predicted masks $M$ and ground truth masks $G$:
\begin{equation}
    \mathcal{E}_{tme} = \frac{1}{T-1} \sum_{t=1}^{T-1} \left| \text{Dice}(M_t, M_{t-1}) - \text{Dice}(G_t, G_{t-1}) \right|
    \label{eq:tme}
\end{equation}
where $T$ denotes the total number of frames.
This metric quantifies the deviation of predicted temporal dynamics from the reference anatomical motion.